\pdfoutput=1
\documentclass[11pt]{article}

\usepackage{emnlp2021}

\usepackage{times}
\usepackage{latexsym}

\usepackage[T1]{fontenc}
\usepackage[utf8]{inputenc}

\usepackage{microtype}

\usepackage{graphicx}
\usepackage{amsmath}
\usepackage{booktabs}
\usepackage{appendix}

\title{Robust Embeddings via Distributions}

\author{Kira A. Selby \\
  University of Waterloo \\
  \texttt{kaselby@uwaterloo.ca} \\\And
  Yinong Wang \\
  University of Waterloo \\
  \texttt{yinong.wang@uwaterloo.ca} \\\AND
  Ruizhe Wang \\
  University of Waterloo \\
  \texttt{r322wang@uwaterloo.ca} \\\And
  Peyman Passban\Thanks{ Work done while Peyman Passban was at Huawei Noah's Ark Lab.} \\
  Huawei Noah's Ark Lab \\
  \texttt{passban.peyman@gmail.com} \\\AND
  Ahmad Rashid \\
  Huawei, Noah's Ark Lab \\
  \texttt{ahmad.rashid@huawei.com} \\\And
  Mehdi Rezagholizadeh \\
  Huawei, Noah's Ark Lab \\
  \texttt{mehdi.rezagholizadeh@huawei.com} \\\AND
  Pascal Poupart \\
  University of Waterloo \\
  \texttt{ppoupart@uwaterloo.ca} \\}

\begin{document}

\maketitle

\begin{abstract}
  Despite recent monumental advances in the field, many Natural Language Processing (NLP) models still struggle to perform adequately on noisy domains. We propose a novel probabilistic embedding-level method to improve the robustness of NLP models. Our method, Robust Embeddings via Distributions (RED), incorporates information from both noisy tokens and surrounding context to obtain distributions over embedding vectors that can express uncertainty in semantic space more fully than any deterministic method. We evaluate our method on a number of downstream tasks using existing state-of-the-art models in the presence of both natural and synthetic noise, and demonstrate a clear improvement over other embedding approaches to robustness from the literature.
\end{abstract}

\section{Introduction}

One major challenge in the area of embeddings is the problem of noise. Embedding schemes using traditional word-based tokenization must use a fixed vocabulary, and cannot handle unknown, or ``out-of-vocabulary" (OOV) words. This can be a major problem when dealing with data mined from social media - an increasingly common domain for NLP tasks - as social media data frequently contains misspellings, slang, or other distortions \cite{DBLP:journals/corr/abs-1711-02173, khayrallah2018impact}. Such noise can cause otherwise highly successful models to completely break down, as demonstrated by \citet{DBLP:journals/corr/abs-1711-02173}.
Traditional pre-processing approaches to noise such as text normalization or spell-checking can be used to mitigate this, but are often outperformed by end to end approaches \cite{malykh2018robust, doval2019towards}. Another approach is to simply train embeddings on noisy corpora, but, as noted by \citet{piktus2019misspelling}, obtaining meaningful representations for all misspelled tokens may require an impractical expansion of the size of the training data. The compromise solution used by many recent models is to use byte-pair encoding (BPE) for tokenization \cite{sennrich-etal-2016-neural}. With BPE, any word can be tokenized and embedded - even words that would otherwise be OOV. The downside to this is that such noisy words will not be embedded as a single token, but rather as multiple subwords or characters. This enables the model to handle noise better than traditional word-based tokenization, but it falls short of a true noise-aware approach.

Instead of using naive preprocessing steps or model-specific strategies to mitigate noise such as BPE or training on noisy corpora, an alternative approach is to train an embedding scheme that is explicitly noise aware. One major advantage of this approach is transferability. Most NLP models use an embedding layer as the first layer of their architecture to convert from discrete tokens to continuous vectors. These embedding layers are typically a sort of black box - a given model architecture can be built on top of a variety of different embedding schemes. As such, a successful noise-aware embedding scheme could be used in place of traditional embedding approaches such as Word2Vec \cite{mikolov2013efficient} or GLoVe  \cite{pennington2014glove} as the first layer of a model without requiring any further alterations to the architecture. Any model built on top of such an embedding scheme would itself become robust to noise, without requiring any further changes to the algorithm.

We propose a probabilistic approach to creating such a robust embedding scheme, which represents each token as a posterior distribution over embedding vectors. Our scheme uses a Bayesian approach, with a prior over possible corrections to the noisy token, and a likelihood function evaluating such corrections in the context of the surrounding sentence. We also introduce an `ensembling' method to turn these distributions into inputs to downstream models in a way that maximizes the available information to the downstream model.Our scheme requires no additional training and can be used in combination with any existing embedding methods. We evaluate our approach on several noisy text classification tasks against a number of simple baselines, as well as other robust vector methods and denoising methods from the literature.

\section{Related work}
\label{sec:related-work}

\subsection*{Robust Word Embeddings } 
Robust Word Vectors (RoVe)~\cite{malykh2018robust} is a morphological context-dependent robust embedding technique targeting typos in the text and is able to deal with open vocabulary. RoVe derives word embeddings in two stages: morphological embedding and context embedding. At the first stage, words are decomposed into beginning (B), middle (M), and end (E) components based on the common prefix, main, suffix word structures. The output embedding of this stage is obtained by aggregating the one-hot representation of the three components. At the second stage, the output morphological embedding from the first stage and both left and right context of the word are fed into an encoder to obtain RoVe embeddings. 

\citet{doval2019towards} proposed a robust word embedding approach based on a modified version of skip-gram by introducing the concept of bridge-words (that is the chain of similar word variants that every two adjacent words are only different in one character, e.g. friend $\rightarrow$ frind $\rightarrow$ freind). They augment existing skip-gram-based methods such as FastText \cite{bojanowski2016enriching} by altering the training process to include bridge-words in addition to each center word. For example, given the sentence ``that's my best friend ever", they train not only to predict ``that's", ``my", ``best" and ``ever" from the center word of ``friend", but also from bridge words such as ``frind" or ``freind". This technique aims to encourage the vector representations for each word to be similar to the vector representations for nearby bridge-words, and thus to encourage the resulting vectors to be more robust to character-based noise.

The MOE approach proposed by Edizel et al \cite{piktus2019misspelling} is modelled on the classic embedding method known as FastText \cite{bojanowski2016enriching}. They use a modified loss function that adds a spelling correction term to improve the robustness of the representations. In addition to a training corpus $T$, MOE also considers a misspelling set $M$ (harvested from social media), consisting of word pairs $(w_m,w_e)$ such that $w_e \in V$ is a correctly spelled word and $w_m$ is a misspelling of that word. Their spelling correction loss term encourages each word vector to be close to the vectors for its misspellings.

\citet{sun2019contextual} deployed pre-trained masked language models such as BERT for denoising text. This technique first corrects noisy inputs before feeding them to downstream tasks. Their method is essentially a form of spelling correction, and lacks the probabilistic framework of our approach. It also looks primarily at the surrounding context to predict the noisy token, and makes little use of the information within the noisy token itself.

\subsection*{Probabilistic Word Embeddings}
Several techniques have been proposed to construct probabilistic embeddings to account for polysemy and homonymy in clean text.  \citet{tian2014probabilistic} extended Word2Vec to account for multiple "prototype" embeddings corresponding to different senses that a word may take.  A latent variable associated with each word is trained by EM to infer the correct sense of each word.  This technique was refined by \citet{liu2015topical} to learn topical latent variables based on the context.  In a different line of work, \citet{vilnis2014word} introduced Gaussian word embeddings to reflect the fact that the precise meaning of a word is uncertain.  The similarity of Gaussian word embeddings can be measured by KL-divergence or Wasserstein distance for greater numerical stability~\cite{sun2018gaussian}. This approach was generalized by replacing Gaussians with more flexible kernels and introducing a mixture to capture disparate senses~\cite{miao2019kernelized}. All those techniques infer distributions over senses only for words in a pre-defined dictionary and cannot deal with out of vocabulary words that arise from noise. Furthermore, it is not always obvious how to build NLP models for downstream tasks on top of those probabilistic embeddings. In contrast, we propose probabilistic embedding techniques that can handle out-of vocabulary words and allow NLP models to be built directly on top without any change.   

\section{Method}
%While these existing approaches all propose interesting solutions to this problem, all of them are fundamentally deterministic. 
The problem of noise in general is one that is intrinsically uncertain. Given a particular misspelled token, there are often multiple possible 'corrections' that one could make to obtain a valid word from the vocabulary - and thus multiple possible semantic roles that a token could perform in a statement. Any single correction to a noisy token is, by necessity, overconfident. A more principled approach is to model possible errors in a probabilistic manner. This allows the embedding to fully capture the range of meanings of the noisy token, and gives the downstream model access to all information rather than only one possible correction.

\subsection{Robust Model}

Our proposal is to represent robust embeddings as distributions over a scheme of ground embeddings. In order to model the distribution for a particular misspelled word, consider how a human responds to seeing a misspelled word. Once the misspelling is identified, we naturally rely on two sources of information to resolve the misspelling: the characters within the word itself (and their similarity to other words we do know), and the surrounding context. Our model thus has two components, which roughly correspond to the Bayesian notions of ``prior" and ``likelihood".

\begin{figure*} [h]
    \centering
    \includegraphics[scale=0.6]{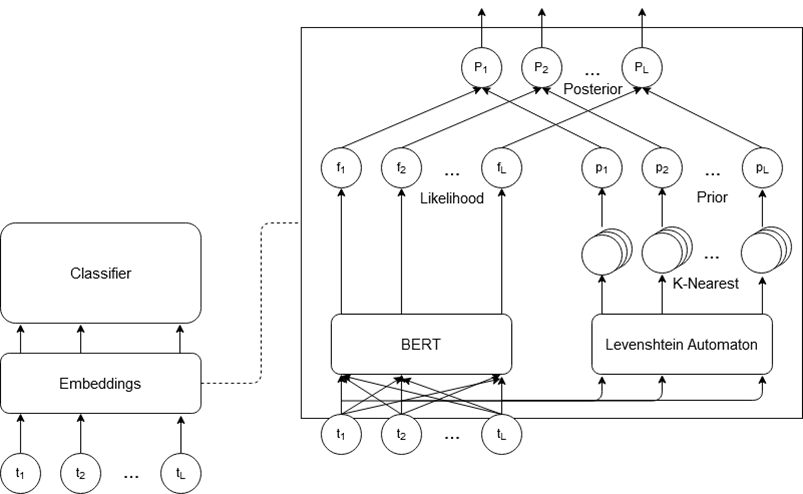}
    \caption{Diagram of the  robust embedding model and how the posterior is computed.}
    \label{fig:model}
\end{figure*}

Suppose we have a sequence of noisy tokens $t_1, ..., t_i, ..., t_l$. In order to produce a robust embedding for $t_i$, we first consider the characters within the token $t_i$ itself. Let $w_i^{(1)}, ..., w_i^{(k)} \in V$ be the $k$ nearest words or tokens in our vocabulary $V$ to the noisy token $t_i$ under a particular string distance metric (e.g. the Levenshtein distance \cite{1966SPhD...10..707L}), with respective distances $d_i^{(1)}, ..., d_i^{(k)}$. Our prior is thus given by:
\begin{equation}
    p\left(w_i^{(j)}\right) = h\left(\vec{d_i}\right)_j
\end{equation}
where $h$ is some normalization function that maps the vector $\vec{d_i} = (d_i^{(1)},...,d_i^{(k)})$ of distances to normalized weights, such that larger distances correspond to smaller weights.
For the purposes of this investigation we use the softmax function, as shown in Eq. \ref{eq:softmax}
\begin{equation}
    h\left(\vec{d}\right)=\text{Softmax}\left(-\frac{\vec{d}}{\tau}\right)
    \label{eq:softmax}
\end{equation}
where $\tau$ is a temperature parameter that controls the concentration of the distribution.

The second component of the distribution is a likelihood function that computes the probability of the surrounding context tokens $t_1, ..., t_{i-1}, t_{i+1},...,t_l$ given a particular center token. The most obvious model for this task is the famous skip-gram formulation of Word2Vec \cite{mikolov2013efficient}, which trains to perform exactly this function. The traditional skip-gram formulation of Word2Vec trains two matrices $W$ and $C$, corresponding to the embeddings of each word in the vocabulary as center and context word respectively. They then predict the likelihood of a particular context-center pair $(c, w)$ occurring based on a softmax of the dot product of $c$ and $w$. We can thus use the $W$ matrix from trained Word2Vec vectors as our ground embeddings for center words, and save the $C$ matrix to be used for the likelihood as in Eq. \ref{eq:sg-lik}:
\begin{equation}
    f_{sg}(C_{t_i}|w) = \prod\limits_{c \in C_{t_i}} \frac{\exp(c \cdot w)}{\sum\limits_{c' \in C} \exp(c' \cdot w)}
    \label{eq:sg-lik}
\end{equation}
where $c \in C_{t_i}$ are the vectors from $C$ corresponding to each of the context words surrounding the original token $t_i$ and $c' \in C$ ranges over all possible context vectors. Given a particular likelihood function, we estimate the posterior for $w$ using Bayes rule, as in Eq. \ref{eq:bayes}:
\begin{equation}
    p(w|C_{t_i}) \propto p(w)f(C_{t_i}|w)
    \label{eq:bayes}
\end{equation}

Unfortunately, skip-gram does not consider word order, and thus it evaluates the probability of each word occurring in the context independently. In practice, this ends up being overly naive, and does not lead to good predictions. While skip-gram is an excellent model for training high-quality intermediate vectors, it is not necessarily the most sophisticated algorithm to actually predict word co-occurrence.
Pre-trained transformer models such as BERT present an alternative to construct a more sophisticated likelihood function. Since BERT is trained using masked language modelling, it naturally computes a likelihood for the masked tokens in an input sequence conditioned on the rest of the sequence.
Given a token $t_i$ at position $i$ in sequence $s$ (with context $C=\{s_j; j \neq i\}$), consider the modified sequence $s'$ with $t_i$ replaced by a mask token. BERT effectively computes:
\begin{align}
    h_{t_i} &= \left[f_{BERT}(s')\right]_i \label{bert1}\\
    f_{BERT}(w|C) &= \frac{\exp{h_{t_i} \cdot w}}{\sum\limits_{v \in V} \exp{h_{t_i} \cdot v}} \label{bert2}
\end{align}

Unfortunately, $f_{BERT}$ gives us a likelihood of $e$ given $C$ instead of $C$ given $e$.  While we cannot use $f_{BERT}$ to compute a posterior with Bayes rule as in Eq.~\ref{eq:bayes}, we can use $f_{BERT}$ as part of a product-of-experts model, as discussed by Hinton in 2002 \cite{hinton2002training}.  We can treat $p(e|w^*)$ and $f_{BERT}(e|C)$ as two expert models that make separate predictions based on different evidence.  Hinton proposed to combine the probabilities of different experts simply by multiplying them and re-normalizing.  Hence we can compute a posterior as follows:
\begin{equation}
    p(e|w^*,C)\propto p(e|w^*) f_{BERT}(e|C)
    \label{eq:poe}
\end{equation}

While this is not precisely a ``posterior" in the Bayesian sense, we continue to use the terms `prior', `likelihood' and `posterior' for convenience.

\begin{figure*} [h]
    \centering
    \includegraphics[scale=0.4]{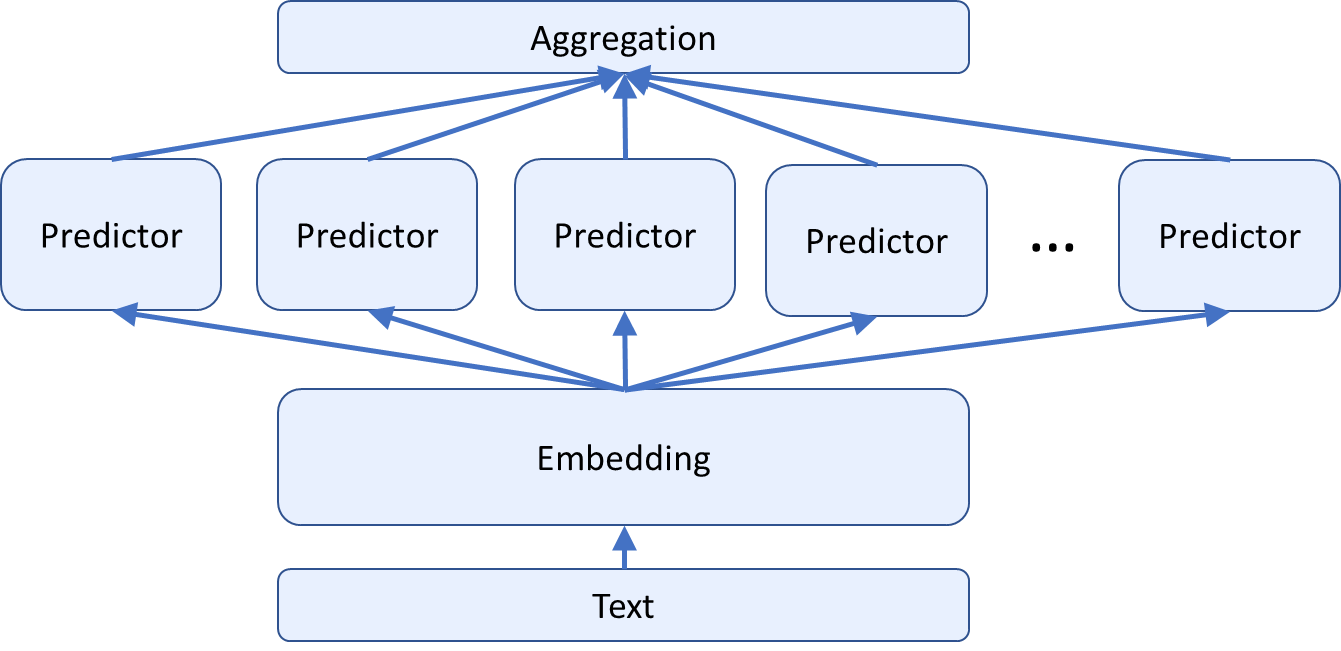}
    \caption{Ensembling model for robust embeddings. K sampled vectors for each token are organized into K sequences. These sequences are each passed through the classifier independently, and aggregated at the end into a single prediction.}
    \label{fig:ensemble}
\end{figure*}

\subsection{Ensembling}
\label{sec:ensemble}
This now results in a method for obtaining a  distribution over the set of ground embeddings which represents the robust embedding of the noisy token $t_i$ given its surrounding context. In practice, however, our resulting embedding vectors will need to be fed into other deep models, which require a single vector as input rather than a distribution. There are several solutions to this. The naive solution is to simply sample a vector from the distribution or take the maximum a posteriori vector. This is perfectly adequate, but does not fully utilize the information contained in the distribution. Instead, we propose to use an ensembling method that more fully captures the full expressiveness of the distribution.

Suppose we have a sequence of noisy tokens $S = \{t_1, ..., t_l\}$ that we wish to embed before passing them into another downstream model (e.g., some sort of text classification model). Each token $t_j$ and its associated context $S_{-j}=\{t_1, ..., t_{j-1}, t_{j+1}, ..., t_l\}$ are passed through the robust embedding model, with the intent of producing robust vector representations of each token respectively. Instead of producing single embedding vectors $e_1, ..., e_l$, the model instead samples $m$ vectors from the posterior distribution for each token, resulting in ensembles $\{ (e_1^{(1)},...,e_1^{(m)}), ..., (e_l^{(1)},...,e_l^{(m)}) \}$. These ensembles of embeddings are then reorganized into an ensemble of embedded sequences $\{ s^{(1)}, ..., s^{(m)} \}$, where $s^{(j)}=\{ e_1^{(j)}, ... e_l^{(j)} \}$. Each sequence is then independently fed into the downstream model, producing $m$ outputs $o_1,...o_m$. These outputs can now be aggregated to produce a final consensus in several ways. The most obvious is to simply average the resulting logits, but other possibilities include taking a majority vote of the predicted classes or adding a shallow feedforward network at the end that could be trained to produce a final, aggregated prediction. We found a simple average of the logits worked best, but refer to the appendices for a more in-depth discussion of other possible aggregation methods.

\begin{table*}
\centering
{\small 
\begin{tabular}{lrrrrrrrrrr}
\toprule
&  \multicolumn{5}{c}{RoBERTa} & \multicolumn{5}{c}{HBMP} \\
\cmidrule(lr){2-6}\cmidrule(lr){7-11}
& & \multicolumn{2}{c}{20\%} & \multicolumn{2}{c}{50\%} & & \multicolumn{2}{c}{20\%} & \multicolumn{2}{c}{50\%}  \\
\cmidrule(lr){3-4}\cmidrule(lr){5-6} \cmidrule(lr){8-9}\cmidrule(lr){10-11}  
Model   & Clean  & Synthetic   & Natural  & Synthetic   & Natural  & Clean   & Synthetic   & Natural  & Synthetic   & Natural  \\
\midrule
\multicolumn{10}{c}{MRPC} \\
\midrule
Naive &0.848 &0.726 &0.727 &0.442 &0.45 &0.764 &0.603 &0.6 &0.403 &0.396 \\
Top-1 &0.843 &0.824 &0.832 &0.766 &0.748 &0.763 &0.711 &0.703 &0.57 &0.56 \\
RED &0.848 &0.832 &\textbf{0.84} &0.793 &0.809 &0.764 &0.737 &0.737 &0.655 &\textbf{0.654} \\
RED-Ens &\textbf{0.859} &\textbf{0.844} &0.838 &\textbf{0.812} &\textbf{0.815} &\textbf{0.771} &\textbf{0.735} &\textbf{0.74} &\textbf{0.658} &0.648 \\
Sun \& Jiang &0.849 &0.837 &0.835 &0.758 &0.768 &0.757 &0.732 &\textbf{0.74} &0.628 &0.649 \\
MOE &x &x &x &x &x &0.73 &0.559 &0.634 &0.42 &0.52 \\
B2V &x &x &x &x &x &0.759 &0.545 &0.583 &0.348 &0.356 \\
\midrule
\multicolumn{10}{c}{QNLI} \\
\midrule
Naive &\textbf{0.903} &0.827 &0.835 &0.705 &0.72 &\textbf{0.809} &0.701 &0.699 &0.614 &0.616 \\
Top-1 &0.899 &0.857 &0.852 &0.804 &0.79 &0.808 &0.751 &0.738 &0.707 &0.686 \\
RED &0.896 &\textbf{0.875} &\textbf{0.874} &\textbf{0.841} &0.832 &0.803 &0.764 &0.762 &0.736 &0.719 \\
RED-Ens &0.887 &0.87 &0.871 &0.838 &\textbf{0.833} &0.800 &\textbf{0.768} &\textbf{0.767} &\textbf{0.743} &\textbf{0.726} \\
Sun \& Jiang &0.885 &0.861 &0.864 &0.803 &0.813 &0.800 &0.756 &0.761 &0.708 &0.705 \\
MOE &x &x &x &x &x &0.772 &0.659 &0.677 &0.583 &0.602 \\
B2V &x &x &x &x &x &0.813 &0.741 &0.752 &0.642 &0.654 \\
\midrule
\multicolumn{10}{c}{QQP} \\
\midrule
Naive &0.879 &0.717 &0.726 &0.450 &0.452 &0.872 &0.721 &0.724 &0.663 &0.659 \\
Top-1 &0.881 &0.754 &0.834 &0.634 &0.634 &0.879 &0.778 &0.777 &0.712 &0.704 \\
RED &\textbf{0.896} &\textbf{0.854} &0.853 &0.807 &0.807 &\textbf{0.889} &0.802 &0.802 &\textbf{0.751} &0.743 \\
RED-Ens &\textbf{0.896} &\textbf{0.854} &\textbf{0.856} &\textbf{0.811} &0.807 &\textbf{0.889} &\textbf{0.806} &0.802 &0.745 &0.743 \\
Sun \& Jiang &0.888 &0.849 & 0.854 &0.795 &\textbf{0.808} &0.860 &0.805 &\textbf{0.812} &0.743 &\textbf{0.759} \\
MOE &x &x &x &x &x &0.875 &0.707 &0.732 &0.639 &0.651 \\
B2V &x &x &x &x &x &0.873 &0.714 &0.735 &0.641 &0.651 \\
\midrule
\multicolumn{10}{c}{SST-2} \\
\midrule
Naive &\textbf{0.934} &0.898 &0.898 &0.816 &0.810 &0.859 &0.772 &0.766 &0.690 &0.664 \\
Top-1 &0.925 &0.893 &0.883 &0.852 &0.838 &0.840 &0.809 &0.800 &0.794 &0.770 \\
RED &0.919 &\textbf{0.911} &0.896 &0.877 &0.861 &0.839 &0.807 &0.805 &0.786 &\textbf{0.787} \\
RED-Ens &0.923 &0.906 &\textbf{0.907} &\textbf{0.880} &\textbf{0.869} &0.837 &\textbf{0.816} &0.811 &\textbf{0.797} &0.781 \\
Sun \& Jiang &0.913 &0.875 &0.882 &0.823 &0.815 &0.834 &0.789 &0.791 &0.755 &0.758 \\
MOE &x &x &x &x &x &0.851 &0.777 &0.793 &0.676 &0.676 \\
B2V &x &x &x &x &x &\textbf{0.861} &0.815 &\textbf{0.819} &0.721 &0.756 \\
\midrule
\multicolumn{10}{c}{CoLA} \\
\midrule
Naive &\textbf{0.561} &0.314 &0.347 &0.163 &0.140 &0.166 &0.155 &0.129 &0.087 &0.071 \\
Top-1 &0.529 &0.401 &0.369 &0.264 &0.241 &0.167 &0.163 &0.128 &0.124 &0.071 \\
RED &0.523 &0.431 &0.408 &0.308 &0.269 &0.150 &0.135 &0.117 &0.137 &\textbf{0.102} \\
RED-Ens &0.551 &\textbf{0.468} &\textbf{0.428} &\textbf{0.333} &\textbf{0.305} &0.166 &\textbf{0.137} &0.133 &\textbf{0.150} &0.091 \\
Sun \& Jiang &0.505 &0.419 &0.402 &0.267 &0.291 &0.140 &0.136 &\textbf{0.135} &0.092 &0.093 \\
MOE &x &x &x &x &x &0.172 &0.097 &0.119 &0.049 &0.055 \\
B2V &x &x &x &x &x &\textbf{0.191} &0.126 &0.112 &0.052 &0.048 \\
\midrule
\multicolumn{10}{c}{MNLI} \\
\midrule
Naive &\textbf{0.864} &0.744 &0.743 &0.591 &0.584 &0.713 &0.552 &0.555 &0.46 &0.456 \\
Top-1 &0.851 &0.763 &0.762 &0.678 &0.663 &\textbf{0.719} &0.619 &0.623 &0.535 &0.529 \\
RED &0.844 &\textbf{0.788} &\textbf{0.784} &\textbf{0.724} &\textbf{0.705} &0.711 &\textbf{0.644} &0.641 &\textbf{0.579} &\textbf{0.57} \\
RED-Ens & 0.838 &0.78 &0.777 &0.72 &0.702 &0.703 &0.639 &0.633 &\textbf{0.579} &0.567 \\
Sun \& Jiang &0.83 &0.771 &0.787 &0.686 &0.703 &0.701 &\textbf{0.644} &\textbf{0.648} &0.568 &0.583 \\
MOE &x &x &x &x &x &0.71 &0.56 &0.588 &0.443 &0.468 \\
B2V &x &x &x &x &x &0.715 &0.593 &0.605 &0.471 &0.478 \\
\midrule
\multicolumn{10}{c}{Average} \\
\midrule
Naive &\textbf{0.831} &0.704 &0.713 &0.528 &0.526 &0.697 &0.584 &0.579 &0.486 &0.477 \\
Top-1 &0.821 &0.749 &0.755 &0.666 &0.652 &0.696 &0.638 &0.628 &0.574 &0.553 \\
RED &0.821 &0.782 &0.776 &0.725 &0.714 &0.693 &0.648 &0.644 &0.607 &\textbf{0.596} \\
RED-Ens &0.823 &\textbf{0.787} &\textbf{0.779} &\textbf{0.732} &\textbf{0.722} &0.694 &\textbf{0.650} &\textbf{0.648} &\textbf{0.612} &0.593 \\
Sun \& Jiang &0.812 &0.769 &0.771 &0.689 &0.700 &0.682 &0.644 &\textbf{0.648} &0.582 &0.591 \\
MOE &x &x &x &x &x &0.685 &0.560 &0.591 &0.468 &0.495 \\
B2V &x &x &x &x &x &\textbf{0.702} &0.589 &0.601 &0.479 &0.490 \\
\bottomrule
\end{tabular}
}
\caption{Results of baselines on GLUE tasks with RoBERTa and HBMP classifiers. An x indicates a method that is not compatible with RoBERTa.}
\label{tab:all-results}
\end{table*}

\section{Experiments}
This method can be evaluated on a variety of downstream tasks. For simplicity, and to focus primarily on the embedding method rather than the details of downstream architectures, we focus on several simple text classification tasks. We use a subset of the tasks from the GLUE text classification benchmark for our experiments \cite{DBLP:journals/corr/abs-1804-07461}, covering tasks such as paraphrase detection, natural language inference, sentiment analysis, and several other forms of text classification. Evaluation is performed on the est set for MRPC, and the development set for all other datasets (as public test sets are not available), with 20\% of the training set withheld as a substitute development set. For each task, we train two different classifiers - a BiLSTM-based approach known as HBMP \cite{talman_yli-jyra_tiedemann_2019}, as well as the state of the art pretrained transformer model "RoBERTa" \cite{DBLP:journals/corr/abs-1907-11692}. Evaluation is performed using a variety of noisy datasets. Since the test set labels are (in general) not publicly available for the GLUE datasets, we use the development sets for evaluation, while withholding a portion of the training set as our new development set. Several different versions of this evaluation set are created: one that is unaltered and has no noise added, as well as versions that have noise injected with either natural noise methods or synthetic noise methods. The noisy corpora are generated by randomly selecting words to have noise injected with a fixed probability of either 20\% or 50\%. Synthetic noise is injected using the method presented in \cite{sun2019contextual} - if a word is selected to have noise injected, there is a 25\% chance each of either deleting a character at a random location, inserting a random character at a random location, swapping the characters at two random locations, or replacing a character at a random location with another random character. Natural noise is injected using a dataset of misspellings harvested from social media (we use the dictionary published by \citet{piktus2019misspelling}). A word selected to have natural noise injected will be replaced by a random misspelling for that word taken from the dataset. Since large, high-quality training sets of noisy data are often unavailable in practice, our experiments are performed by training the downstream model on clean training data only, with noise applied only during evaluation. See Section \ref{sec:analysis} for further discussion of the effects of training on noisy data.

We compare our method against several baselines. First, we compare against the relatively naive baselines of either a) simply using the downstream model with no additional attempt to improve the robustness, or b) correcting each out-of-vocabulary word to the nearest word under the Levenshtein distance. We also compare against several baselines from the literature. For the experiments using HBMP, we compare against a variety of baselines, including scratch-trained vectors such as Bridge2Vec~\cite{doval2019towards} and MOE~\cite{piktus2019misspelling}, and the denoising method of \citet{sun2019contextual}. Unfortunately, it is not practical to compare against many of these methods in the case of RoBERTa, as replacing the input embedding layer of RoBERTa would require retraining the entire model. Our model does not suffer from this issue since it can output embeddings from any predefined embedding space and therefore can be inserted below any pretrained model such as RoBERTa without requiring retraining.

\begin{table*} [t]
\centering
{\small
\begin{tabular}{lr}
\toprule
Method  &   Text \\
\midrule
Original Sequence   & someone is in a bowling alley. \\
Noisy Sequence  & oneone is yn a bowlind aley . \\
Top-1 Levenshtein   & someone is yan a bowling alley.\\
Sun \& Jiang & none is in a blind alley. \\
RED    & someone is in a bowling alley. \\
\midrule
Original Sequence: & ``the great thing is to keep calm . "" julius groaned ." \\
Noisy Sequence & ``the great thing si ot dkeep calj . "" julius groaned." \\
Top1 Levenshtein & the great thing si ot keep call . " julius groaned \\
Sun \& Jiang & the great thing i got keep call . " julius grinned \\
RED & the great thing is to keep calm . " julius groaned \\

\bottomrule
\end{tabular}
}
\caption{Comparison against other baselines as a denoising method on example noisy sequences.}
\label{tab:examples}
\end{table*}

\section{Results}
Experiments were performed using Nvidia T4 and P100 GPUs. %provided by the Vector Institute. 
FastText vectors were used as the ground embeddings for the Naive method, our method, and all spellchecking/denoising-based methods for the HBMP classifier. Implementation details for the HBMP classifier were taken from their publicly-available code\footnote{https://github.com/Helsinki-NLP/HBMP}, with 400-dimensional hidden layers and default hyperparameter settings. Implementation details for the RoBERTa models and GLUE tasks were taken from the HuggingFace repository\footnote{https://github.com/huggingface/transformers} \cite{wolf-etal-2020-transformers}, with default hyperparameter settings. Each experiment was performed three times, and the results were averaged across the trials. Implementations of baselines were taken from publically available source code where possible, and otherwise were reimplemented based on descriptions of the methods. Further details on the experimental setup and baseline implementations are contained in the appendices.

For the robust model, the softmax temperature parameter in the prior was taken to be $\tau=0.1$, with the top $k=10$ nearest words under the Levenshtein distance considered, and the ensemble models used $m=10$ samples each. We use the likelihood function described in equations \ref{bert1} and \ref{bert2}, with the standard pretrained BERT model published by HuggingFace.

\subsection{RoBERTa}
Table \ref{tab:all-results} shows results for each baseline and noise setting across the various tasks, as well as the average results across all tasks. Our technique, RED (Robust Embeddings via Distributions) with ensembling performs the best on average in all noise settings except for the completely clean case. The base RED model without ensembling performs slightly worse in all categories while still outperforming all other baselines on average. The benefits of ensembling seem highly dependent on the specific task, with the ensemble method being clearly superior on CoLA, MRPC and SST-2, relatively similar to the base method on QNLI, and QQP, and slightly worse on MNLI. Aside from the case of completely clean data, our models score highest on all categories across all datasets except QQP, where the model of Sun \& Jiang slightly outperforms them on one of the nine categories. 
The slight decrease in performance compared to the naive baseline on clean data is expected, due to the nature of our method. Since there should be few to no errors or ambiguous tokens, the cases in which our model is more likely to make a correction are those where words are out of vocabulary for other reasons, such as names, acronyms, slang, rare words, etc. Trying to model these tokens as possible misspellings of other in-vocabulary tokens is unlikely to help disambiguate the meaning of their surrounding sentence, and thus it is not unexpected that the performance on completely clean data may suffer slightly.

\subsection{HBMP}
Again, the ensemble model performs best on average across most categories, with the highest or tied for the highest score on every noise setting except the completely clean case and the 50\% natural noise category with clean training. While RED and RED-Ensemble still perform the best, the gap is much smaller than in the RoBERTa case. On some tasks such as SST-2, other baselines perform better across certain categories - particularly for the natural noise cases with noisy training. We hypothesize that since this classifier is less powerful than RoBERTa, it is less able to fully leverage the information provided by our model.

\section{Analysis}
\label{sec:analysis}
\subsection{Examples}

\begin{table*} 
\centering
{\small
\begin{tabular}{lp{10cm}}
\toprule
Method  &   Text \\
\midrule
Original Sequence & we did not study the reasons for these deviations specifically , but they likely result from the context in which federal cios operate . \\
Noisy Sequence & we di nor study the reasons for these deviations specifically, but thew likely reault from the context in which fideral ctos operate. \\
Ensemble Samples & we dig nor study the reasons for these deviation specifically but they likely result from the context in which federal cops operates \\
& we di nor study the reasons for these depictions specifically but the likely result from the context in which federal cops operate \\
& we did nor study the reasons for these deviation specifically but they likely result from the context in which federal cops operate \\
& we did nor study the reasons for these deviation specifically but the likely result from the context in which federal actors operate \\
& we di nor study the reasons for these deviation specifically but the likely result from the context in which federal actors operate \\
\bottomrule
\end{tabular}
}
\caption{Examples}
\label{tab:examples2}
\end{table*}

\begin{table*} [h]
\centering
{\small
\begin{tabular}{lrrrr}
\toprule
Candidate  &   Log-Prior & Log-Likelihood & Log-Posterior & Probability \\
\midrule
cops    & -2.99573  & -8.44108  & -11.4368  & 0.474132\\
actors  & -2.99573  & -8.44204  & -11.4378  & 0.473674\\
crows   & -2.99573  & -11.3638  & -14.3595  & 0.0255029\\
stops   & -2.99573  & -11.9221  & -14.9179  & 0.0145912\\
atoms   & -2.99573  & -13.5329  & -16.5286  & 0.0029145\\
\bottomrule
\end{tabular}
}
\caption{Distribution for the second-to-last word of the sequence shown in Table~\ref{tab:examples2}}
\label{tab:distribution}
\end{table*}

%\midrule
%Original Sequence   & a group of kids lay on a colorful structure. \\
%Noisy Sequence  & a group of lids olay n a colorful structure. \\
%Top-1 Levenshtein   & a group of lids play n a colorful structure.\\
%Ours    & a group of lids lay on a colorful structure. \\
%Ours (with ensemble)    & a group of lids lay on a colorful structure. \\
%& a group of lads lay in a colorful structure .\\
%& a group of lids lay on a colorful structure .\\
%& a group of lids lay on a colorful structure .\\
%& a group of kids lay in a colorful structure . \\
%Original Sequence   & a man reads the paper in a bar with green lighting. \\
%Noisy Sequence  & a man reads the paler in a aar ith green lighing.\\
%Top-1 Levenshtein   & a man reads the paler in a amar with green sighing.\\
%Ours    & a man reads the paper in a car it green lighting. \\
%Ours (with ensemble)  &  a man reads the paper in a car with green lighting.\\
%& a man reads the paler in a ear it green lighting .\\
%& a man reads the paler in a bar with green lighting.\\
%& a man reads the paler in a car with green lighting.\\
%& a man reads the paper in a war with green lighting.\\
%\midrule

In order to illustrate the workings of the model, we will now highlight some examples of the model applied to some sample noisy sequences from the data. We can thus compare the RED model to some of the other baselines from the viewpoint of text correction, and show some examples of the resulting posterior distributions and the effect of the ensemble method in the context of concrete examples.

In Table~\ref{tab:examples}, we compare our model against the top1-Levenshtein and Sun \& Jiang baselines on some difficult noisy sequences taken from the data. Even in cases where the sequences are very noisy, our model can often generate reasonable corrections where the other models cannot. In some ways, our model can be viewed as a form of interpolation between the top-1 Levenshtein approach - which only looks at the characters within the word - and the approach of Sun \& Jiang, which considers primarily the context. By more fully utilizing both sources of information, our model can successfully tackle cases that neither of those baselines can adequately model.

In order to study how the model makes predictions, consider the example shown in Table~\ref{tab:examples2}. The "true" correction for the second-to-last token is somewhat unclear, even for a human reader. The distribution produced by the model (shown in Table~\ref{tab:distribution}) is concentrated around two possibilities "cops" and "actors" - both of which could be reasonable corrections in the context. By using the ensemble approach, we can take a mixture of samples that show multiple possibilities for ambiguous tokens such as these, and thus provide the downstream model with more information about the possible semantic content of the sentence. This gives the model more power and flexibility than a deterministic "spellchecker", such as the other baselines we compare against.

\subsection{Noisy Training}
Since our model is one that does not require training and can easily be applied at evaluation time, a natural question is whether it would perform worse relative to other baselines in the case where noisy training data is available. In order to test this, we ran experiments with the same settings wherein noise was injected to the training data as well as the test data. Noise was injected in the same fashion as the previous experiments. Average results across all tasks are shown in Table~\ref{tab:noisy-results}; full results by task are included in the appendices. The ensemble model is once again equal or superior to all other baselines across both the RoBERTa and HBMP classifiers.

\begin{table*} [t]
\centering
{\small 
\begin{tabular}{lrrrrrrrr}
\toprule
&  \multicolumn{4}{c}{RoBERTa} & \multicolumn{4}{c}{HBMP} \\
\cmidrule(lr){2-5}\cmidrule(lr){6-9}
& \multicolumn{2}{c}{20\%} & \multicolumn{2}{c}{50\%} & \multicolumn{2}{c}{20\%} & \multicolumn{2}{c}{50\%}  \\
\cmidrule(lr){2-3}\cmidrule(lr){4-5} \cmidrule(lr){6-7}\cmidrule(lr){8-9}  
Model  & Synthetic   & Natural  & Synthetic   & Natural   & Synthetic   & Natural  & Synthetic   & Natural  \\
\midrule
Naive &0.763 &0.748 &0.634 &0.610 &0.621 &0.611 &0.573 &0.562 \\
Top-1 &0.750 &0.773 &0.715 &0.719 &0.656 &\textbf{0.663} &0.630 &0.623 \\
RED &0.779 &0.780 &0.751 &\textbf{0.739} &0.661 &\textbf{0.663} &0.634 &0.628 \\
RED-Ensemble &\textbf{0.794} &\textbf{0.783} &\textbf{0.761} &\textbf{0.739} &\textbf{0.665} &\textbf{0.663} &\textbf{0.639} &\textbf{0.629} \\
Sun \& Jiang &0.774 &0.773 &0.712 &0.719 &0.653 &0.656 &0.622 &0.623 \\
MOE &x &x &x &x &0.616 &0.616 &0.520 &0.575 \\
B2V &x &x &x &x &0.643 &0.645 &0.558 &0.591 \\

\bottomrule
\end{tabular}
}
\caption{Average results across all datasets for experiments with noisy training data.}
\label{tab:noisy-results}
\end{table*}

\subsection{CO2 Emission Related to Experiments}
In accordance with the suggestions presented in \cite{lacoste2019quantifying}, we report the approximate computation time and carbon emissions of experiments associated with this work.

Experiments were conducted using a private infrastructure, which has a carbon efficiency of 0.04 kgCO$_2$eq/kWh. A cumulative of 50000 hours of computation was performed on hardware of type T4 (TDP of 70W).

Total emissions are estimated to be 140 kgCO$_2$eq of which 0 percent were directly offset. 140 kgCO$_2$eq were manually offset through \href{www.less.ca}{Less Emissions Inc}.

Estimations were conducted using the \href{https://mlco2.github.io/impact#compute}{MachineLearning Impact calculator} presented in \cite{lacoste2019quantifying}.

\section{Conclusion}
In conclusion, we propose a novel approach to robustness at the embedding level that is probabilistic and transferable. Our method can be used as a foundation for any other NLP model in order to make that model more robust, without requiring any changes to the architecture. Our approach is probabilistic, and thus more fully expresses the uncertainty in meaning within a noisy sequence than other similar models. Using the ensembling approach we propose, the downstream model can have access to all (or at least many) possible meanings of the noisy tokens, and can be trained end to end to decide for itself how to use this information - rather than making overconfident corrections at the preprocessing level. We show superior results on a range of noisy tasks, using both synthetic and natural noise.

\bibliography{anthology,custom}
\bibliographystyle{acl_natbib}

\appendix
\appendixpage
\section{Aggregation Methods}
\label{sec:agg}
As discussed in Section \ref{sec:ensemble}, the ensemble model produces $m$ embedded sequences, $\{ s^{(1)}, ..., s^{(m)} \}$, where $s^{(j)}=\{ e_1^{(j)}, ... e_l^{(j)} \}$. Each sequence is then independently fed into the downstream model, producing $m$ outputs $o_1,...o_m$. For a classification-based downstream task with $c$ classes, these outputs will often take the form of logits, resulting in $m \times c$ scalar outputs. As discussed earlier, the simplest ways to aggregate these into a single prediction is to either a) average the logits across the ensemble, then take the argmax or b) take a majority vote of the argmaxed logits of the individual outputs. We found the former method to be most effective, and it is used throughout the paper. Many other methods of performing this aggregation exist however. One simple possibility to add additional flexibility to the aggregation is to add a simple multilayer perceptron as a final layer, taking the $m \times c$ logits as input and producing a single set of logits as output. This could be trained alongside the downstream model. The difficulty with this method is that the $m$ outputs are each constructed from samples from the same distribution, and as such are interchangeable in order. This can cause significant difficulties in optimization for a naive MLP, and we found empirically that the convergence was poor. Improving this area of the model is a possible site for future work, using either an MLP with some selection of statistics about the sampled outputs as features (e.g. mean, standard deviation, min and max, etc...), or possibly a bespoke architecture designed to account for the symmetry in the inputs.

\section{Experiments and Baselines}
Independent noisy versions of the data were constructed for each trial, which were each then used to train each of the two classifiers on the appropriate task, with each robust method used in turn to mitigate the effects of the noise. Basic preprocessing was applied to the noisy data before passing it through the robust models (e.g. lowercasing, stripping some characters, etc...). Experiments were performed using the glue finetuning script from HuggingFace, with all training parameters left as default (save for the CoLA case with the HBMP classifier, where 20 epochs were used instead of the default 3 due to the very small size of the dataset). For the HBMP case, FastText vectors were used as a ground embedding layer for the methods that did not proscribe their own vectors.

Baselines were evaluated using existing source code where possible (Bridge2Vec), and otherwise were reimplemented following as closely to the authors' original descriptions as possible (MOE, Sun \& Jiang). A value of $\alpha=0.05$ was used for the MOE training.

\section{Noisy Training Results}
Results of experiments with noisy training data across all tasks and noise levels are shown in Table \ref{tab:noisy-full}. As before, RED and RED-Ensemble show superior performance across almost all tasks and noise levels save for the completely clean cases.

\begin{table*}
\centering
{\small 
\begin{tabular}{lrrrrrrrr}
\toprule
&  \multicolumn{4}{c}{RoBERTa} & \multicolumn{4}{c}{HBMP} \\
\cmidrule(lr){2-5}\cmidrule(lr){6-9}
& \multicolumn{2}{c}{20\%} & \multicolumn{2}{c}{50\%} & \multicolumn{2}{c}{20\%} & \multicolumn{2}{c}{50\%}  \\
\cmidrule(lr){2-3}\cmidrule(lr){4-5} \cmidrule(lr){6-7}\cmidrule(lr){8-9}  
Model   & Synthetic   & Natural  & Synthetic   & Natural    & Synthetic   & Natural  & Synthetic   & Natural  \\
\midrule
\multicolumn{9}{c}{MRPC} \\
\midrule
Naive &0.782 &0.798 &0.696 &0.712 &0.699 &0.707 &0.662 &0.681 \\
Top-1 &0.836 &0.832 &0.806 &0.807 &0.74 &0.743 &0.708 &0.71 \\
RED &\textbf{0.848} &\textbf{0.843} &0.817 &0.821 &0.745 &\textbf{0.761} &0.725 &0.726 \\
RED-Ensemble &0.844 &0.842 &\textbf{0.839} &\textbf{0.831} &\textbf{0.751} &0.757 &\textbf{0.739} &\textbf{0.732} \\
Sun \& Jiang  &0.833 &0.836 &0.798 &0.805 &0.746 &0.754 &0.718 &0.729 \\
MOE &x &x &x &x &0.68 &0.692 &0.657 &0.672 \\
B2V &x &x &x &x &0.726 &0.721 &0.685 &0.685 \\
\midrule
\multicolumn{9}{c}{QNLI} \\
\midrule
Naive &0.854 &0.862 &0.777 &0.782 &0.727 &0.734 &0.651 &0.643 \\
Top-1 &0.875 &0.869 &0.847 &0.843 &0.778 &0.779 &0.761 &0.754 \\
RED &0.878 &\textbf{0.88} &\textbf{0.861} &0.853 &\textbf{0.782} &0.779 &0.760 &\textbf{0.753} \\
RED-Ensemble &\textbf{0.881} &0.875 &0.859 &\textbf{0.854} &\textbf{0.782} &\textbf{0.780} &\textbf{0.763} &0.743 \\
Sun \& Jiang  &0.866 &0.873 &0.822 &0.824 &0.770 &0.773 &0.741 &0.742 \\
MOE &x &x &x &x &0.664 &0.676 &0.594 &0.615 \\
B2V &x &x &x &x &0.762 &0.764 &0.692 &0.702 \\
\midrule
\multicolumn{9}{c}{QQP} \\
\midrule
Naive &0.839 &0.841 &0.802 &0.804 &0.790 &0.757 &0.795 &0.746 \\
Top-1 &0.877 &0.876 &0.855 &0.855 &0.830 &0.831 &0.791 &0.801 \\
RED &\textbf{0.878} &0.875 &\textbf{0.860} &\textbf{0.856} &0.833 &0.832 &0.804 &0.804 \\
RED-Ensemble &0.876 &\textbf{0.878} &\textbf{0.860} &0.819 &\textbf{0.834} &\textbf{0.835} &\textbf{0.807} &\textbf{0.810} \\
Sun \& Jiang  &0.865 &0.867 &0.834 &0.842 &0.830 &0.831 &0.793 &0.801 \\
MOE &x &x &x &x &0.810 &0.815 &0.788 &0.778 \\
B2V &x &x &x &x &0.82 &0.821 & &0.784 \\
\midrule
\multicolumn{9}{c}{SST-2} \\
\midrule
Naive &0.900 &0.902 &0.857 &0.843 &0.793 &0.790 &0.738 &0.727 \\
Top-1 &0.904 &0.899 &0.872 &0.862 &0.819 &0.824 &0.815 &0.796 \\
RED &0.904 &0.899 &\textbf{0.890} &0.866 &0.819 &0.816 &0.810 &0.799 \\
RED-Ensemble &\textbf{0.906} &\textbf{0.904} &0.884 &\textbf{0.877} &0.820 &\textbf{0.823} &\textbf{0.814} &\textbf{0.810} \\
Sun \& Jiang  &0.879 &0.878 &0.828 &0.832 &0.798 &0.814 &0.772 &0.775 \\
MOE &x &x &x &x &0.822 &0.808 &0.761 &0.767 \\
B2V &x &x &x &x &\textbf{0.828} &0.822 &0.78 &0.774 \\
\midrule
\multicolumn{9}{c}{CoLA} \\
\midrule
Naive &0.397 &0.354 &0.022 &0.095 &0.127 &0.085 &0.074 &0.060 \\
Top-1 &0.360 &0.350 &0.223 &0.193 &0.104 &0.133 &0.086 &0.059 \\
RED &0.347 &0.367 &0.299 &0.268 &0.115 &0.121 &0.072 &0.063 \\
RED-Ensemble &\textbf{0.451} &\textbf{0.393} &\textbf{0.336} &\textbf{0.291} &\textbf{0.135} &\textbf{0.122} &\textbf{0.079} &0.055 \\
Sun \& Jiang  &0.404 &0.382 &0.256 &0.265 &0.112 &0.094 &0.097 &\textbf{0.073} \\
MOE &x &x &x &x &0.082 &0.069 &0.029 &0.049 \\
B2V &x &x &x &x &0.074 &0.092 &0.053 &0.013 \\
\midrule
\multicolumn{8}{c}{MNLI} \\
\midrule
Naive &0.806 &0.732 &0.65 &0.427 &0.588 &0.596 &0.519 &0.516 \\
Top-1 &0.646 &0.809 &0.688 &0.756 &0.665 &0.665 &0.62 &0.618 \\
RED &\textbf{0.816} &\textbf{0.814} &0.777 &\textbf{0.767} &\textbf{0.671} &\textbf{0.668} &\textbf{0.635} &\textbf{0.625} \\
RED-Ensemble &0.809 &0.805 &\textbf{0.79} &0.764 &0.667 &0.662 &0.63 &0.622 \\
Sun \& Jiang &0.796 &0.801 &0.733 &0.745 &0.663 &\textbf{0.668} &0.611 &0.62 \\
MOE &x &x &x &x &0.636 &0.638 &0.557 &0.571 \\
B2V &x &x &x &x &0.649 &0.651 &0.581 &0.59 \\
\bottomrule
\end{tabular}
}
\caption{Results of baselines on GLUE tasks with noisy training with both RoBERTa and HBMP classifiers. An x indicates a method that is not compatible with RoBERTa.}
\label{tab:noisy-full}
\end{table*}

\end{document}